\def\year{2019}\relax
\documentclass[letterpaper]{article} 
\usepackage{aaai19}  
\usepackage{times}  
\usepackage{helvet}  
\usepackage{courier}  
\usepackage{url}  
\usepackage{graphicx}  

\usepackage{epsfig}
\usepackage{rotating,tabularx}
\usepackage{amsmath}

\usepackage{amssymb}
\usepackage{subcaption}
\usepackage{mathtools}
\usepackage{algorithm}
\usepackage{algpseudocode}
\usepackage{comment}

\newcommand{\keypoint}[1]{\noindent\textbf{#1}\quad}

\newcommand{\cut}[1]{}

\makeatletter
\def\BState{\State\hskip-\ALG@thistlm}
\makeatother

\usepackage{multirow}
\usepackage[table,xcdraw]{xcolor}

\usepackage{footnote}

\begin{document}

\title{Disjoint Label Space Transfer Learning with Common Factorised Space}

\author{Xiaobin Chang$^1$, Yongxin Yang$^2$, Tao Xiang$^1$, Timothy M. Hospedales$^2$\\
Queen Mary University of London$^1$, The University of Edinburgh$^2$\\
{\tt\small x.chang@qmul.ac.uk yongxin.yang@ed.ac.uk t.xiang@qmul.ac.uk t.hospedales@ed.ac.uk}
}

\maketitle
\begin{abstract}
In this paper, a unified approach is presented to transfer learning that addresses several source and target domain label-space and annotation assumptions with a single model. It is particularly effective in handling a challenging case, where source and target label-spaces are \emph{disjoint}, and outperforms alternatives in both unsupervised and semi-supervised settings. 
The key ingredient is a common representation termed \emph{Common Factorised Space}. It is shared between source and target domains, and trained with an unsupervised factorisation loss and a graph-based loss. 
With a wide range of experiments, we demonstrate the flexibility, relevance and efficacy of our method, both in the challenging cases with disjoint label spaces, and in the more conventional cases such as unsupervised domain adaptation, where the source and target domains share the same label-sets.

\end{abstract}

\section{Introduction}\label{Sec:Into}

\begin{figure}[t]
\centering
\includegraphics[width=0.85\columnwidth]{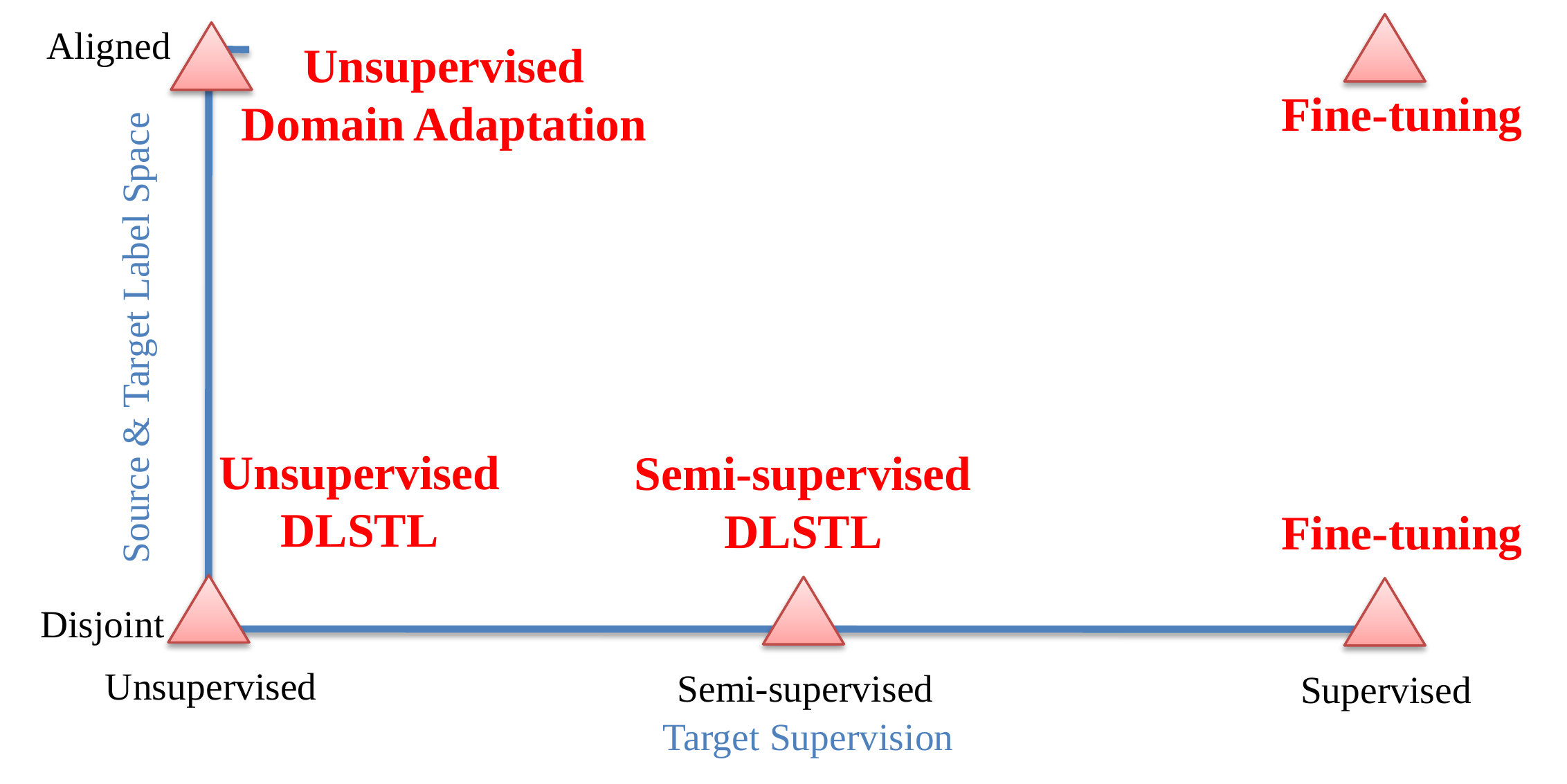}
\caption{Schematic of various transfer learning problems on two criteria: the relation between source and target label space, and the amount of target problem supervision.}
\label{fig:tl_illustrate}
\end{figure}

Deep learning methods are now widely used in diverse applications. However, their efficacy is largely contingent on a large amount of labelled data in the target task and domain of interest. This issue continues to motivate intense interest in cross-task and cross-domain knowledge transfer. A wide range of transfer learning settings are considered which differ in whether the label spaces of source and target domains are overlapped (i.e., aligned or disjoint), as well as the amount of supervision/labelled training samples available in the target domain (see Figure~\ref{fig:tl_illustrate}). The standard practice of \emph{fine-tuning} \cite{yosinski2014transferable} treats a pre-trained source model  as a good initialisation for training a target problem model. It is adopted when the label spaces of both domains are either aligned or disjoint, but always requires a significant amount of labelled data from the target, albeit less than learning from scratch. 
Another popular problem is the unsupervised domain adaptation (UDA), where knowledge is transferred from a labelled source domain to an unlabelled target domain \cite{adversarial_feat_cvpr17,ganin2016domain,cao2018unsupervised}. UDA makes the simplifying assumption that the label space of source and target domains are the same, and focuses on narrowing the distribution gap between source and target domains without any labelled samples from the target.

An important but less-studied transfer learning problem setting is one where the source and target domains are with \emph{disjoint} label spaces,
recently highlighted by \cite{label_eff_open_da_2017}. In these problems, which we term Disjoint Label Space Transfer Learning (DLSTL), there are both a domain shift between source and target, as well as a new set of target classes to recognise with few (semi-supervised case) or no labelled (unsupervised case) sample per category. Thus, two main challenges exist simultaneously. On one hand, there is few or no target label to drive the adaptation. On the other hand, no clear path is provided to transfer source supervision to target domain due to the disjoint label spaces.
As an example, consider object recognition in two cameras (domains) where the object categories (label-space) are different in each camera, and one source camera has dense labels, while the target camera has data but few or no labels.
The traditional fine-tuning \cite{yosinski2014transferable} and multi-domain training \cite{rebuff2017mdl} can address the supervised (few label) DLSTL variant, but break down if the labels are very few, and cannot exploit unlabelled data in the target camera, i.e., semi-supervised learning. 
Meanwhile UDA approaches~\cite{ganin2016domain} based on distribution alignment are ineffective since the label-spaces are disjoint and feature distributions thus should not be indistinguishable.
One approach that has the potential to handle DLSTL under both unsupervised and semi-supervised settings is based on modelling attributes, which can serve as a bridge across domains for better transferring the discriminative power~\cite{chen2015deep,gebru2017fine,attr_label_transfer_2018wang}. Source and target data can be aligned within the attribute space, in order to alleviate the impacts of disjoint label space in DLSTL problems. Nevertheless, attribute can be expensive to acquire which prevents it form being widely applicable.

In this paper, a novel transfer learning model is proposed, which focuses on handling the most challenging setting, \emph{unsupervised DLSTL} but is applicable to other settings including  semi-supervised DLSTL and UDA.
The model, termed common factorised space model (CFSM), is 
developed based on the simple idea that recognition should be performable in a shared latent factor space for both domains where each factor can be interpreted as latent attribute \cite{fu2013latentAttrib,farhadi2012attributeDiscovery}.
In order to automatically discover such discriminative latent factors and align them for transferring knowledge across datasets/domains, 
our inductive bias is that input samples from \emph{both} domains should generate \emph{low-entropy} codes in this common space, i.e., near binary-codes \cite{salakhutdinov2009semantic,zhu2016deep}. This is a weaker assumption than distribution matching, but  does provide a criterion that can be optimised to align the two domains in the absence of common label space and/or labelled target domain training samples. Specifically, both domains should be explainable in terms of the same set of discriminative latent factors with high certainty.
As a result, discriminative information from the source domain can be more effectively transferred to the target through this common factorised space. 
To implement this model in a neural network architecture, a common factorised space (CFS) layer is inserted between the feature output layer (the penultimate layer) and the classification layer (the final layer). This layer is shared between both domains and thus forms a common space.
An unsupervised factorisation loss is then derived and applied on such common space which serves the purpose of optimising low-entropy criterion for discriminative latent factors discovery.

Somewhat uniquely, cross-domain knowledge transfer of the proposed CFSM occurs at a relatively high layer (i.e., CFS layer). Particularly when the target domain problem is a retrieval one, it is important that this knowledge is propagated down from CFS to feature extraction for effective knowledge transfer. To assist this process we define a novel graph Laplacian-based loss - which builds a graph in the higher-level CFS, and regularises the lower-level network feature output to have matching similarity structure. i.e., that inter-sample similarity structure in the shared latent factor space should be reflected in earlier feature extraction. 
This top-down regularisation is opposite to the use of Laplacian regularisation in existing works \cite{belkin2006laprlsSSL,yang2017graph} which are bottom-up, i.e., graph from lower-level regularises the higher-level features. This unique design is due to the fact that, although both spaces (CFS and feature) are latent, the former is closer to supervisions (e.g., from the labelled source data) and more aligned thanks to the factorisation loss, and thus more discriminative and `trustworthy'. 

Contributions of the paper are as follows: 
1. A unified approach to transfer learning is proposed. It can be applied to different transfer learning settings but is particularly attractive in handling the most challenging  setting of unsupervised DLSTL. This setting is under-studied with the latest efforts focus on the easier semi-supervised DLSTL setting~\cite{label_eff_open_da_2017} with partially labelled target data. Several topical applications in computer vision such as person re-identification (Re-ID) and sketch-based image retrieval (SBIR) can be interpreted as unsupervised DLSTL which reveals its vital research and application values.
2. We propose a deep neural network based model, called common factorised space model (CFSM), that provides the first simple yet effective method for unsupervised DLSTL; it can be easily extended to semi-supervised DLSTL as well as conventional UDA problems.
3. A novel graph Laplacian-based loss is proposed to better exploit the more aligned and discriminative supervision from higher-level to improve deep feature learning.
Finally, comprehensive experiments on various transfer learning settings, from UDA to DLSTL, are conducted. CFSM achieves state-of-the-art results on both unsupervised and semi-supervised DLSTL problems and performs competitively in standard UDA. The effectiveness and flexibility of the proposed model on transfer learning problems are thus demonstrated.

\section{Related Work}\label{Sec:Relate_W}

\subsection{Transfer Learning}
Transfer learning (TL) aims to transfer knowledge from one domain/task to improve performance on the another \cite{pan2010survey}. The most widely used TL technique for deep networks is fine-tuning \cite{yosinski2014transferable,chen2018lstd,ren2015faster}. Instead of training a target network from scratch, its weights are initialised by a pre-trained model from another  task such as ImageNet \cite{deng2009imagenet} classification. 
While fine-tuning reduces label requirement compared to learning the target problem from scratch, it is prone to over-fitting if target labels are very few  \cite{yosinski2014transferable}. Therefore, it is ineffective for very sparsely supervised DLSTL, and not applicable to unsupervised DLSTL. 
Moreover, vanilla TL does not exploit available unlabelled samples for the target problem (i.e. semi-supervised TL). The most related method to ours is \cite{label_eff_open_da_2017} which does exploit both unlabelled and few labelled data, i.e., semi-supervised DLSTL. However like other TL methods, it does not generalise to the unsupervised DLSTL setting where no target annotations are available.

Another popular setting,  unsupervised domain adaptation (UDA)  focuses on transferring the source supervision to the unlabelled target domain in order to obtain a model that performs well on the latter data. The typical assumption of UDA is that both domains share the same label space. Existing methods alleviate the domain gap by either minimising the distribution discrepancy \cite{cao2018unsupervised,deep_coral} or making the dataset representations indistinguishable by adversarial learning \cite{adversarial_feat_cvpr17,ganin2016domain}. Once the domain gap is eliminated, a classifier trained on source-domain labels can be applied to the target data directly. However, distribution matching is inappropriate in the disjoint label space setting.
\textcolor{black}{Open set domain adaptation (OSDA) \cite{busto2017open} generalises UDA by allowing target domain to have some novel categories in addtion to the shared ones. It focuses on identifying shared categories and aligning those. DLSTL is a more general problem setting than OSDA, since there is no assumption of any shared categories.}
Related to our approach that exploits a common factorisation space to discover shared latent factors/attributes, semantic attributes have been used to improve domain adaptation performance \cite{su2016deep}, for example by enabling new types of self-training \cite{chen2015deep,attr_label_transfer_2018wang} and consistency losses \cite{gebru2017fine}. However these methods require the attribute definition and  annotation, at least in the source domain. In contrast, no expensive attribute annotation is required in our model.

\subsection{Deep Binary Representation Learning}

The use of binary codes for hashing with deep networks goes back to \cite{salakhutdinov2009semantic}. 
In computer vision, hashing layers were inserted between feature- and classification-layers to provide a hashing code \cite{lin2015deep,zhu2016deep}. To produce a binary representation for fast retrieval, a threshold is applied on the sigmoid activated hashing layer \cite{lin2015deep}. 
Our method is similar in working with a near-binary penultimate layer. However there are several key differences: First, 
our CFS serves a very different purpose to a hash code. We focus on TL to a new domain with new label-space, and the role of our CFS is to provide a representation with which  different domains can be more aligned for knowledge transfer, rather than for efficient retrieval. In contrast, existing hashing methods follow the conventional supervised learning paradigm within a single domain. Second, the proposed CFS is only near-binary due to a low-entropy loss, rather than sacrificing representation power for an exactly binary code.

\subsection{Semi-supervised Learning}
Graph-based regularisation is popular for semi-supervised learning (SSL) which uses both labelled and unlabelled data to achieve better performance than learning with labelled data only \cite{zhu2006semi,belkin2006laprlsSSL}. 
In SSL,
graph based regularisation is applied to regularise model predictions to respect the feature-space manifold \cite{yue2017semi,nadler2009semi,belkin2006laprlsSSL}. 
Moreover, exploiting the graph from lower-level to regularise higher-level features is widely adopted in other scenarios, e.g., unsupervised learning~\cite{jia2015laplacian,yang2017graph}.
Due to the source$\to$target knowledge transfer, the more `trustworthy' layer in our method is the penultimate CFS layer, as it is closer to the supervision, rather than the feature space layer. 
Therefore our regularisation is applied to encourage the feature-extractor to learn representations that respect the CFS manifold shared by both domains, i.e., the regularisation direction is opposite to that in existing models.

Entropy loss for unlabelled data is another widely used SSL regulariser \cite{zhu2006semi,long2016unsupervised}. It is applied at the classification layer in problems where the unlabelled and labelled data share the same label-space -- and reflects the inductive bias that a classification boundary should not cut through the dense unlabelled data regions. Its typical use is on softmax classifier outputs where it encourages a classifier to pick a single label. In contrast we use entropy-loss to solve DLSTL problems by applying it  element-wise on our intermediate CFS layer in order to weakly align domains by encouraging them to share a near-binary representation.

\section{Methodology}\label{Sec:Method}

\paragraph{Definition and notation}

For Disjoint Label Space Transfer Learning (DLSTL), there is a source (labelled) domain $\mathcal{S}$ and a target (unlabelled or partially labelled) domain $\mathcal{T}$\footnote{The proposed model can be easily extended to deal with multiple source and target domains}. The key characteristic of DLSTL is the disjoint label space assumption, i.e., the source $\mathcal{Y}_\mathcal{S}$ and target $\mathcal{Y}_\mathcal{T}$ label spaces are potentially disjoint: $\mathcal{Y}_\mathcal{S} \cap \mathcal{Y}_\mathcal{T} = \varnothing$. Instances from source/target domains are denoted  $X_\mathcal{S}$ and $X_\mathcal{T}$ respectively. The combined inputs $\{X_\mathcal{S}, X_\mathcal{T}\}$ are denoted as $X$. To present our model, we stick mainly to the most challenging unsupervised DLSTL setting where target labels are totally absent. The easier cases, e.g., semi-supervised DLSTL and UDA, can then be handled with minor modifications.

\subsection{Model Architecture}
\begin{figure}[t]
\centering    
\includegraphics[width=0.85\columnwidth]{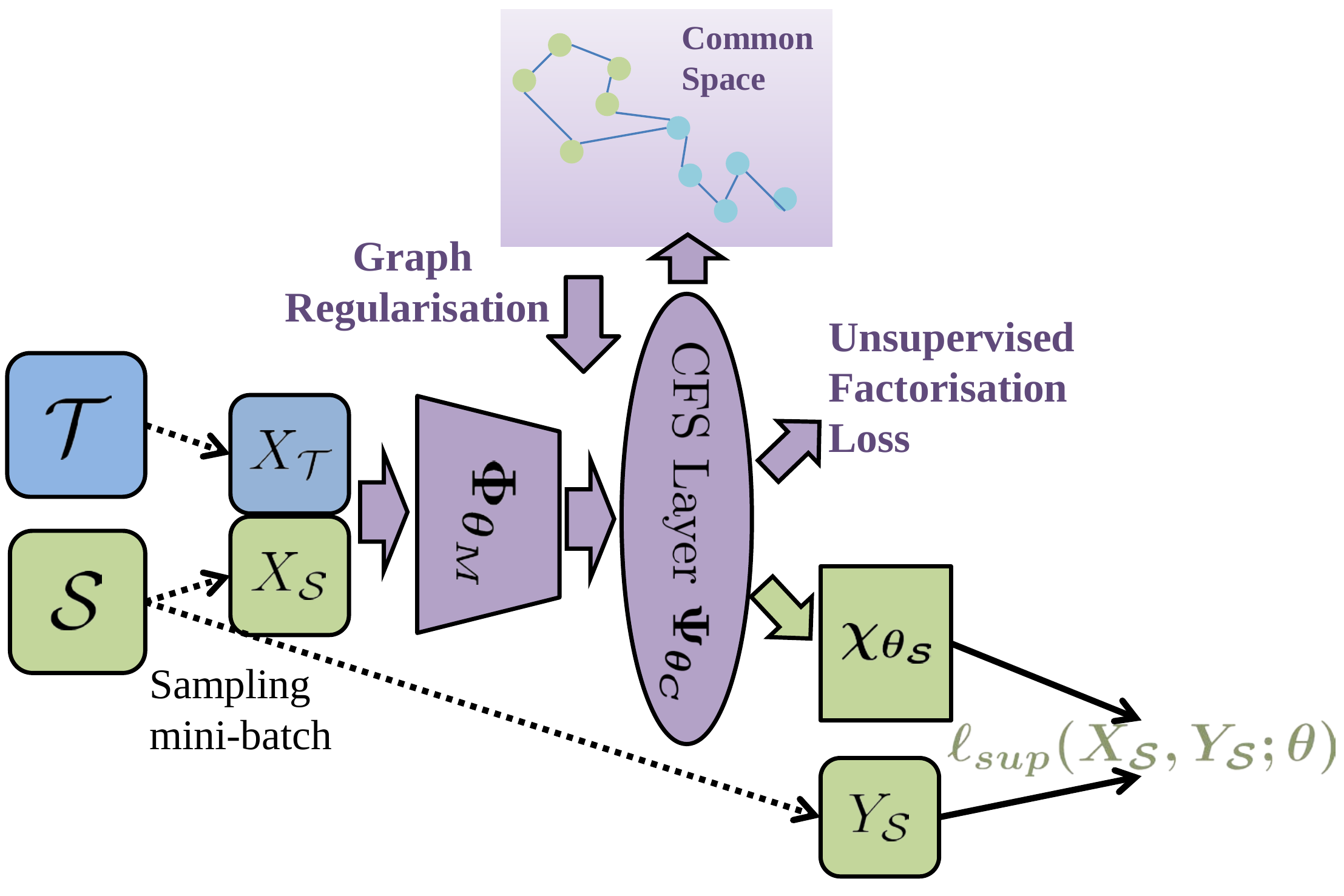}
\caption{The proposed model architecture is illustrated. Different colours corresponding to different data streams. Green indicates source data. Blue is used for target data. Purple means joint data from both source and target domains.}
\label{fig:model_arch_illus}
\end{figure}

The proposed model architecture consists of three modules, a feature extractor $F=\Phi_{\theta_M}(X)$ that can be any deep neural network and is shared between all domains. This is followed by a fully connected layer and sigmoid activation $\sigma$, which define the Common Factorised Space (CFS) layer. This provides a representation of dimension $d_C$, $F_C=\Psi_{\theta_C}(\cdot)={\sigma}(W\Phi_{\theta_M}(\cdot)+b)$. Recall that the goal of CFS is to learn a latent factor (low-entropy) representation for both source and target domains. The sigmoid activation means that the layer's scale is $F_C\in(0,1)^{d_C}$, so activations near 0 or 1 can be interpreted as the corresponding latent factor being present or absent. To encourage a near-binary representation, unsupervised factorisation loss is applied.
For the labelled source domain only, the pre-activated $F_C$ are then classified by softmax classifier $\chi_{\theta_\mathcal{S}}$ with cross-entropy loss. 
The overall architecture is illustrated in Figure~\ref{fig:model_arch_illus}.

\subsection{Regularised Model Optimisation}
The parameters of the proposed CFSM are $\theta \coloneqq \{\theta_{M}, \theta_{C}, \theta_{S}\}$ including parameters of the feature extractor $\theta_M$, CFS layer $\theta_C$ and source classifier $\theta_{S}$. The training procedure can be formulated as a maximum-a-posterior (MAP) learning given labelled source $\{X_\mathcal{S}, Y_\mathcal{S}\}$ and unlabelled target data $X_\mathcal{T}$,
\begin{equation}\label{eq:MAP}
	\hat{\theta} = \underset{\theta}{\operatorname{argmax}}~ p(\theta | X_\mathcal{S}, Y_\mathcal{S}, X_\mathcal{T}),
\end{equation}
where $p(\theta | X_\mathcal{S}, Y_\mathcal{S}, X_\mathcal{T})$ is the posterior of model parameter $\theta$ given data $X_\mathcal{S}, Y_\mathcal{S}, X_\mathcal{T}$.
This can be rewritten as
\begin{equation}\label{eq:propto_MAP}
\begin{split}
	p(\theta | X_\mathcal{S}, Y_\mathcal{S}, X_\mathcal{T})
	\propto & p(\theta, X_\mathcal{S}, Y_\mathcal{S}, X_\mathcal{T}) \\
	\propto &p(Y_\mathcal{S} | X_\mathcal{S}, X_\mathcal{T}, \theta) p(\theta | X_\mathcal{S}, X_\mathcal{T}). 
\end{split}
\end{equation}

So the optimisation in Eq.~\ref{eq:MAP} is equivalently
\begin{equation}\label{eq:optm_propto_MAP}
	\hat{\theta} = \underset{\theta}{\operatorname{argmax}}~ p(Y_\mathcal{S} | X_\mathcal{S}, \theta) p(\theta | X).
\end{equation}
The first term $p(Y_\mathcal{S} | X_\mathcal{S}, \theta)$ in Eq.~\ref{eq:optm_propto_MAP} represents the likelihood of source labels w.r.t. $\theta$. Optimising this term is a conventional supervised learning task with a loss denoted $\ell_{sup}(X_\mathcal{S}, Y_\mathcal{S}; \theta)$. 

The second term $p(\theta | X)$ in Eq.~\ref{eq:optm_propto_MAP} is a prior depending on the input data $X$ of both source and target datasets. From an optimisation perspective, this is the regulariser that will play the key role in solving unsupervised DLSTL problems since it requires no labels. Given the model architecture, it can be further decomposed as:

\begin{equation}\label{eq:prior_decomp_C_M}
\begin{split}
   p(\theta | X) = & p(\theta_M, \theta_C | X) \\
	= & p(\theta_C | \theta_M, X) p(\theta_M | X),
\end{split}
\end{equation}
where $\theta_{\mathcal{S}}$ is excluded since no supervision is used. Specifically, the first term $p(\theta_C | \theta_M, X)$ serves as the regulariser on the CFS layer while the second term $p(\theta_M | X)$ regularises the deep feature extractor $\Phi_{\theta_M}$.

\paragraph{Low-Entropy Regularisation: Unsupervised Adaptation}

We first discuss how to define the prior $p(\theta_C | \theta_M, X)$ regulariser for CFS layer. The sigmoid activated outputs $F_C$ from CFS layer $\Psi_{\theta_C}$ can be interpreted as multi-label predictions on latent factors. The uncertainty measure for label prediction can be defined by using its entropy
\begin{equation}\label{eq:entropy_loss}
\begin{split}
	h(\theta_C | \theta_M, X) = &- \sum_{i=1}^{N} <F_{C,i}, \log(F_{C,i})>\\
	                                         =& - \sum_{i=1}^{N} <\Psi_{\theta_C}(\mathbf{x}_i), \log(\Psi_{\theta_C}(\mathbf{x}_i))>
\end{split}
\end{equation}
where $F_{C,i}$ denotes the common factor representation $\Psi_{\theta_C}(\mathbf{x}_i)$ of instance $\mathbf{x}_i \in X$. This is applied on both source and target data, so $N$  is the number of instances in both datasets. $\log(\cdot)$ is applied element-wise, and $<\cdot,\ \cdot>$ is vector inner product.
According to the low-uncertainty criterion \cite{carlucci2017autodial}, optimising the prior term $p(\theta_C | \theta_M, X)$ can be achieved by minimising this uncertainty measure. Eq.~\ref{eq:entropy_loss} is thus the regulariser corresponding to the prior $p(\theta_C | \theta_M, X)$. 
Specifically, this loss biases the representation $F_C$ to contain more certain predictions, e.g., closer to 0 or 1 for each discovered latent factor. Therefore, we denote it as unsupervised factorisation loss.

\textcolor{black}{In summary, the low-entropy regulariser on CFS is built upon the assumption that the two domains share a set of latent attributes and that if a source classifier is well adapted to the target, then the presence/absence of these attributes should be certain for each instance. Therefore, it essentially generalises the low-uncertainty principle (widely used in existing unsupervised and semi-supervised learning literature) to the disjoint label space setting.}

\paragraph{Graph Regularisation: Robust Feature Learning}
The second prior in Eq.~\ref{eq:prior_decomp_C_M} is $p(\theta_M | X)$ which acts as the regulariser for the feature extractor $\Phi_{\theta_M}$. The unique property of our setup so far is that the knowledge transfer into the target domain is via the CFS layer; therefore we are interested in ensuring that the feature extractor network extracts features whose similarity structure reflects that of the latent factors in the CFS layer. Unlike conventional graph Laplacian losses that regularise higher-level features with a graph built on lower-level features \cite{belkin2006laprlsSSL,zhu2006semi}, we do the reverse and regularise the feature extractor $\Phi_{\theta_M}$ to reflect the similarity structure in $F_C$.  This is particularly important for applications where the target problem is retrieval, because we use deep features $F=\Phi_{\theta_M}(\cdot)$ as an image representation.

The proposed graph loss is expressed as
\begin{equation}\label{eq:graph_loss}
	\operatorname{Tr}(F^{T} \Delta_{F_C} F),
\end{equation}
where  $\Delta_{F_C}$ is the graph Laplacian~\cite{cai2011graph} built on the common space features $F_C$.

\keypoint{Summary}
We unify the proposed model architecture $\theta := \{\theta_M, \theta_C, \theta_{D_\mathcal{S}}\}$ with source $\{X_\mathcal{S}, Y_\mathcal{S}\}$ and target  $\{X_\mathcal{T}\}$ data for unsupervised DLSTL problems from an maximum-a-posterior (MAP) perspective. This decomposes into a standard supervised term $p(Y_\mathcal{S}|X_\mathcal{S},\theta)$ (source data only) and data-driven priors for the CFS layer and feature extraction module. 
They correspond to supervised loss $\ell_{sup}(X_\mathcal{S},Y_\mathcal{S};\theta)$, unsupervised factorisation loss (Eq.~\ref{eq:entropy_loss}) and the graph loss (Eq.~\ref{eq:graph_loss}) respectively.
Taking all terms into account, the final optimisation objective of Eq.~\ref{eq:optm_propto_MAP} is 
\begin{equation}\label{eq:final_loss}
\begin{split}
	L(\theta) = &\textcolor{black}{\ell_{sup}(X_\mathcal{S},Y_\mathcal{S};\theta) + \beta_M Tr(F^{T} \Delta_{F_C} F)}\\
	& -\beta_C \frac{1}{N} \sum_{i=1}^{N} <F_{C,i}, \log(F_{C,i})>. \\
\end{split}
\end{equation}
where $\beta_C$ and $\beta_M$ are balancing hyper-parameters.
\textcolor{black}{In order to select $\beta_C$ and $\beta_M$, the model is first run by setting all weights to 1;  after the first few iterations, we check the values of each loss. We then set the two hyper-parameters to rescale the losses to a similar range so that all three terms contribute approximately equally to the training.}

\keypoint{Mini-batch organisation} Convolutional Neural Networks (CNNs) are usually trained with SGD mini-batch optimisation, but Eq.~\ref{eq:final_loss} is expressed in a full-batch fashion. Converting Eq.~\ref{eq:final_loss} to mini-batch optimisation is straightforward. 
However, it is worth mentioning the mini-batch scheduling: each mini-batch contains samples from both source and target domains. The supervised loss is applied only to source samples with corresponding supervision, the entropy and graph losses are applied to both, and the graph is built per-mini-batch. In this work, the number of source and target samples are equally balanced in a mini-batch.




\section{Experiments}\label{Sec:Exp}
The proposed model is evaluated on progressively more challenging problems.   
First, we evaluate CFSM on unsupervised domain adaptation (UDA).
Second, different DLSTL settings are considered, including semi-supervised DLSTL classification and unsupervised DLSTL retrieval. CFSM handles all these  scenarios with minor modifications. The effectiveness CFSM is demonstrated by its superior performance compared to the existing work. 
Finally insight is provided through ablation study and visualisation analysis.

\subsection{Unsupervised Domain Adaptation: SVHN-MNIST}

\keypoint{Dataset and Settings} We evaluate the UDA setting from \cite{ganin2016domain} where SVHN~\cite{svhn_data} is the labelled source dataset and MNIST~\cite{mnist_data} is the unlabelled target. For fair comparison we use an identical feature extractor network to  \cite{label_eff_open_da_2017}. Our CFSM is pre-trained on the source dataset with cross-entropy supervision and $d_C=50$, followed by joint training on source and target with our regularisers as in Eq.~\ref{eq:final_loss}. Since the label-space is shared in UDA, we also apply entropy loss on the softmax classification of the target \cite{long2016unsupervised}. We set $\beta_M=0.001$ and $\beta_C=0.01$. 

\keypoint{Results} We compare our method with two baselines. Source only: Supervised training on the source and directly apply to target data. Joint FT: Model is initialised with source pre-train, and fine-tuning on both domains with supervised loss for source and semi-supervised entropy loss for target. We also compare several deep UDA methods including Gradient Reversal~\cite{ganin2016domain}, Domain Confusion~\cite{tzeng2015simultaneousTransfer}, ADDA~\cite{adversarial_feat_cvpr17}, Label Efficient Transfer (LET)~\cite{label_eff_open_da_2017}, Asym. tri-training~\cite{saito2017asymmetric} and Res-para~\cite{rozantsev2018residual}.


As shown in Table~\ref{tab:UDA_svhn_mnist}, CFSM boosts the performance on both baselines with clear margin ($25.5\%$ and $9.3\%$ vs. Source only and Joint FT respectively). Moreover, it is $5.5\%$ higher than LET \cite{label_eff_open_da_2017}, the nearest competitor and only alternative that \emph{also} addresses the DLSTL setting.

\begin{table}[t]
\centering
\small
\begin{tabular}{llc}
\hline
Method           &       &Accuracy \\ \hline
Domain confusion & ICCV'15 & 68.1    \\
Grad. reversal   & JMLR'16 & 73.9    \\
ADDA             & CVPR'17 & 76.0    \\
LET	             & NIPS'17 & 81.0    \\ 
Res-para         & CVPR'18 & 84.7    \\ 
Asym. tri-training     & ICML'17 & 85.0    \\ \hline
Source only      &  & 61.0    \\
Joint FT         &  & 77.2    \\
CFSM             &  & {\bf 86.5}    \\ \hline
\end{tabular}
\caption{Unsupervised domain adaptation results. Classification accuracy ($\%$) on SVHN$\to$MNIST transfer.}
\label{tab:UDA_svhn_mnist}
\end{table}

\subsection{Semi-supervised DLSTL: Digit Recognition}\label{Sec:DLSTL_category}
\keypoint{Dataset and Settings} We follow the semi-supervised DLSTL recognition experiment of \cite{label_eff_open_da_2017} where again two digit  datasets, SVHN and MNIST, are used. Images of digits 0 to 4 from SVHN are fully labelled as source data while images of digits 5 to 9 from MNIST are target data. The target dataset has sparse labels ($k$ labels per class) and unlabelled images available. Thus we also add a classifier $\chi_{\theta_\mathcal{T}}$ after the CFS layer $\Psi_{\theta_C}$ for the target categories. 

The feature extractor architecture $\Phi_{\theta_M}$ is exactly the same as in \cite{label_eff_open_da_2017} for fair comparison. \cut{The labelled and unlabelled data are subject to no other losses than those in  Eq.~\ref{eq:final_loss}.} 
We pre-train CFSM on source data as initialisation, and then train it with both source and target data using only loss in Eq.~\ref{eq:final_loss}. We set $d_C=10, \beta_M=\beta_C=0.01$. The learning rate is $0.001$ and the Adam~\cite{kingma2014adam} optimiser is used.

\keypoint{Results}
The results for several degrees of target label sparsity $k = 2,3,4,5$ (corresponding to $10,15,20,25$ labelled samples, or $0.034\%,0.050\%,0.066\%,0.086\%$ of total target training data respectively), are reported in Table~\ref{tab:digit_svhn04_mnist59}. Results are averaged over ten random splits as in \cite{label_eff_open_da_2017}. Besides the FT matching nets~\cite{vinyals2016matching} and state-of-the-art LET results from \cite{label_eff_open_da_2017}, we run two baselines: Train Target: Training CFSM architecture from scratch with partially labelled target data only, and FT Target: The standard pre-train/fine-tune pipeline, i.e., pre-train on the labelled source, and fine-tune on the labelled target samples only.

\begin{table*}[!htb]
\centering
\small
\begin{tabular}{cc|c|c|c|c}
\hline
             & & $k=2$  & $k=3$ & $k=4$ & $k=5$ \\ \hline
Train Target & & $ 66.5 \pm 1.7 $  &  $ 77.2 \pm 1.1 $  &  $ 83.0 \pm 0.9 $  &  $ 88.3 \pm 1.1 $  \\
FT Target    & & $ 69.8 \pm 1.6 $  &  $ 79.1 \pm 1.2 $  &  $ 84.5 \pm 0.8$  &  $ 89.3 \pm 0.9$  \\
FT matching nets & NIPS'16 & $ 64.5 \pm 1.9 $ & $ 75.5 \pm 2.4 $ & $ 79.3 \pm 1.3 $ & $ 82.7 \pm 1.1 $ \\
LET  & NIPS'17 &  $91.7 \pm 0.7$     &  $93.6 \pm 0.6$    &   $94.2 \pm 0.6$    &   $95.0 \pm 0.4$    \\ \hline
CFSM          & &   $ \mathbf{93.5 \pm 0.5} $  &  $ \mathbf{94.8 \pm 0.5} $  &  $ \mathbf{95.5 \pm 0.3} $  &  $ \mathbf{96.7 \pm 0.2} $  \\ \hline
\end{tabular}
\caption{Semi-supervised DLSTL image categorisation results (\%), with mean classification accuracy and standard error for SVHN (0-4) $\to$ MNIST (5-9).}
\label{tab:digit_svhn04_mnist59}
\end{table*}

As shown in Table~\ref{tab:digit_svhn04_mnist59}, the performances of baseline models are significantly lower than LET and the proposed CFSM. The Train Target baseline performs poorly as it is hard to achieve good performance with few target samples and no knowledge transfer from source. The Fine-Tune Target baseline performs poorly as the annotation here is too sparse for effective fine-tuning on the target problem. 
Fine-Tune matching nets follows the $5$-way $(k-1)$-shot learning with sparsely labelled target data only, but no improvement is shown over the other baselines.
Our proposed CFSM consistently outperforms the state-of-the-art LET alternative. For example, under the most challenging setting ($k=2$), CFSM is $1.8\%$ higher than LET on mean accuracy and $0.2\%$ lower on standard error.

\subsection{Unsupervised DLSTL: ReID and SBIR}
\subsubsection{ReID}
The person re-identification (ReID) problem is to match person detections across camera views. Annotating person image identities in every camera in a camera network for training supervised models is infeasible. This motivates the topical unsupervised  Re-ID problem of adapting a  Re-ID model trained on one dataset with annotation to a new dataset without annotation. Although they are evaluated with retrieval metrics, contemporary Re-ID models are trained using identity prediction (classification) losses. This means that unsupervised Re-ID fits the \emph{unsupervised} DLSTL setting, as the label-spaces (person identities) are different in different Re-ID datasets, and the target dataset has no labels.

We adopt two highly contested large-scale benchmarks for unsupervised person Re-ID: Market~\cite{zheng2015scalable} and Duke~\cite{zheng2017unlabeled}. ImageNet pre-trained Resnet50~\cite{resnet} is used as the feature extractor $\Phi_{\theta_M}$. Cross-entropy loss with label smoothing and triplet loss are used for the source domain as supervised learning objectives. We set $d_C=2048,\beta_M=2.0,\beta_C=0.01$. Adam optimiser is used with learning rate $3.5e^{-4}$. We treat each dataset in turn as source/target and perform unsupervised transfer from one to the other. 
 Rank 1 (R1) accuracy and mean Average Precision (mAP) results on the target datasets are used as evaluation metrics. 
 
\begin{table}[t]
\centering
\small
\begin{tabular}{ll|cc|cc}
\hline
\multicolumn{1}{c}{} & \multicolumn{1}{c|}{} & \multicolumn{2}{c|}{M2D} & \multicolumn{2}{c}{D2M} \\ \hline
model      &           & R1          & mAP         & R1           & mAP         \\ \hline
UMDL       &  CVPR'16     &     18.5    &    7.3     &     34.5     &    12.4     \\
PTGAN      &  CVPR'18     &     27.4        &     -        &  38.6            &    -         \\
PUL        &  arXiv'17    &     30.0    &    16.4     &     45.5     &   20.5      \\
CAMEL      &  ICCV'17     &     -    &      -   &     54.5     &   26.3      \\
TJ-AIDL    &  CVPR'18         & 44.3        & 23.0        & 58.2         & 26.5        \\
SPGAN      &  CVPR'18         & 46.4        & 26.2        & 57.7         & 26.7        \\ 
MMFA       &  BMVC'18   &      45.3       &   24.7          &   56.7           &  27.4           \\ \hline
CFSM       &            & {\bf 49.8}        & {\bf  27.3}        & {\bf 61.2}         & {\bf 28.3}        \\ \hline
\end{tabular}
\caption{Unsupervised transfer for person Re-ID ($\%$). M2D indicates Market as  source dataset and Duke as target, vice versa. Target Dataset Performance is reported.}
\label{tab:Unsup_ReID}
\end{table}

In Table~\ref{tab:Unsup_ReID}, We show that our method outperforms the state-of-the-art alternatives purpose-designed for unsupervised person Re-ID: UMDL~\cite{peng2016unsupervised}, PTGAN~\cite{wei2017_trans_GAN}, PUL~\cite{fan2017unsupervised_reid}, CAMEL~\cite{yu2017cross}, TJ-AIDL~\cite{attr_label_transfer_2018wang}, SPGAN~\cite{deng2017image-image} and MMFA~\cite{lin2018multi}.
Note that TJ-AIDL and MMFA exploit attribute labels to help alignment and adaptation. The proposed method automatically discovers latent factors with no additional annotation. However, CFSM improves at least $3.0\%$ over TJ-AIDL and MMFA on the R1 accuracy of both settings.

\subsubsection{FG-SBIR}
Fine-grained Sketch Based Image Retrieval (SBIR) focuses on matching a sketch with its corresponding photo \cite{sketchy2016}. As demonstrated in \cite{sketchy2016}, object category labels play an important role in retrieval performance, so existing studies make a closed world assumption, i.e., all testing categories overlap with training categories. However, if deploying SBIR in a real application such as e-commerce \cite{yu2016sketchShoe}, one would like to train the SBIR system once on some source object categories, and then deploy it to provide sketch-based image retrieval of new categories without annotating new data and re-training for the target object category. Unsupervised adaptation to new categories
without sketch-photo pairing labels is therefore  another example of the unsupervised DLSTL problem. Comparing to  Re-ID, where instances are person images in different camera views, instances in SBIR are either photos or hand-drawn sketches of objects.

There are $125$ object classes in the Sketchy dataset~\cite{sketchy2016}. We randomly split $75$ classes as a labelled source domain and use the remaining $50$ classes to define an unlabelled target domain with disjoint label space. ImageNet pre-trained Inception-V3 \cite{szegedy2016rethinking} is used as the feature extractor $\Phi_{\theta_M}$. Cross-entropy and triplet loss are used for source supervision.  We set $d_C=512,\beta_M=10^{-3},\beta_C=0.1$. Adam optimiser with learning rate $10^{-4}$ is used. As a baseline, Source Only is the direct transfer alternative that uses the same architecture but trains on the source labelled data only, and is applied directly to the target without adaptation.
The retrieval performance on unseen classes (tar. cls.) are reported. Results are averaged over $10$ random splits.
As shown in Table~\ref{tab:DLSTL_Sketchy}, the proposed CFSM improves the retrieval accuracy on unseen cases by $2.48\%$.

\begin{table}[htb]
\centering
\small
\begin{tabular}{c|c|c}
\hline
          & Source only        & CFSM                 \\ \hline
tar. cls. & $23.74 \pm 0.24$   & $\mathbf{26.22 \pm 0.25}$    \\ \hline
\end{tabular}
\caption{SBIR: Sketch-photo retrieval results ($\%$). Averaged Rank 1 accuracy and standard error.}
\label{tab:DLSTL_Sketchy}
\end{table}




\subsection{Further Analysis}
\keypoint{Ablation study}  Unsupervised person Re-ID is chosen as the main benchmark for an ablation study. Firstly because it is a challenging and realistic large-scale problem in the unsupervised DLSTL setting, and secondly because it provides a bidirectional evaluation for more comprehensive analysis.

The following ablated variants are proposed and compared with the full CFSM. Source Only: The proposed architecture is learned with source data and supervised losses only. Source+Regs: The regularisers, unsupervised factorisation and graph losses can be added with source dataset only. CFSM$-$Graph: Our method without the proposed graph loss. CFSM+ClassicGraph: Replacing our proposed graph loss with a conventional Laplacian graph (i.e., graphs constructed in lower-level feature space extracted by $\Phi_{\theta_M}$ to regularise the proposed CFS). AE: Other regularisers such as feature reconstruction as in autoencoder (AE) is used to provide the prior term $p(\theta|X)$. We reconstruct the deep features $F$  using the outputs of CFS layer as hidden representations. In this case both source and target data are used and the reconstruction error provides the regularisation loss. 

The results are shown in Table~\ref{tab:reid_ablation}. Firstly, by comparing the variants that use source data only (Source Only and Source+Regs) with the joint training methods, we find they are consistently inferior. This illustrates that it is crucial to leverage target domain data for adaptation. 
Secondly, CFSM and its variants consistently achieve better results than AE, illustrating that our unsupervised factorisation loss and graph losses provide better regularisation for cross-domain/cross-task adaptation. The effectiveness of our graph loss is illustrated by two comparisons: (1) CFSM$-$Graph is worse than CFSM, showing the contribution of the graph loss; and (2) replacing our graph loss with the conventional Laplacian graph loss (CFSM+ClassicGraph) shows worse results than ours, justifying our choice of regularisation direction.
Finally, we note that applying our regularisers to the source only (Source+Regs) still improves the performance slightly on target dataset vs Source Only. This shows that training with these regularisers has a small benefit to representation transferability even without adaptation.


\begin{table}[t]
\centering
\small
\begin{tabular}{c|cc|cc}
\hline
\multicolumn{1}{c|}{} & \multicolumn{2}{c|}{M2D} & \multicolumn{2}{c}{D2M} \\ \hline
model                 & R1          & mAP         & R1           & mAP         \\ \hline
Source Only              & 39.2       & 20.2        & 54.4         & 23.0        \\
Source+Regs            & 41.6       & 21.2        & 55.8         & 24.0        \\
AE                     & 43.6       & 22.8        & 56.4         & 24.9        \\ \hline
CFSM$-$Graph            & 46.8        & 25.6        & 60.0         & 27.6        \\ 
CFSM+ClassicGraph              & 47.4        & 26.1        & 59.0         & 27.0        \\ 
CFSM                    & {\bf 49.8}        & {\bf 27.3}        & {\bf 61.2}         & {\bf 28.3}        \\ \hline
\end{tabular}
\caption{Ablation study on unsupervised person Re-ID benchmarks. Target dataset performance ($\%$) is reported.}
\label{tab:reid_ablation}
\end{table}



\begin{figure}[t]
\centering    
\includegraphics[width=0.95\columnwidth]{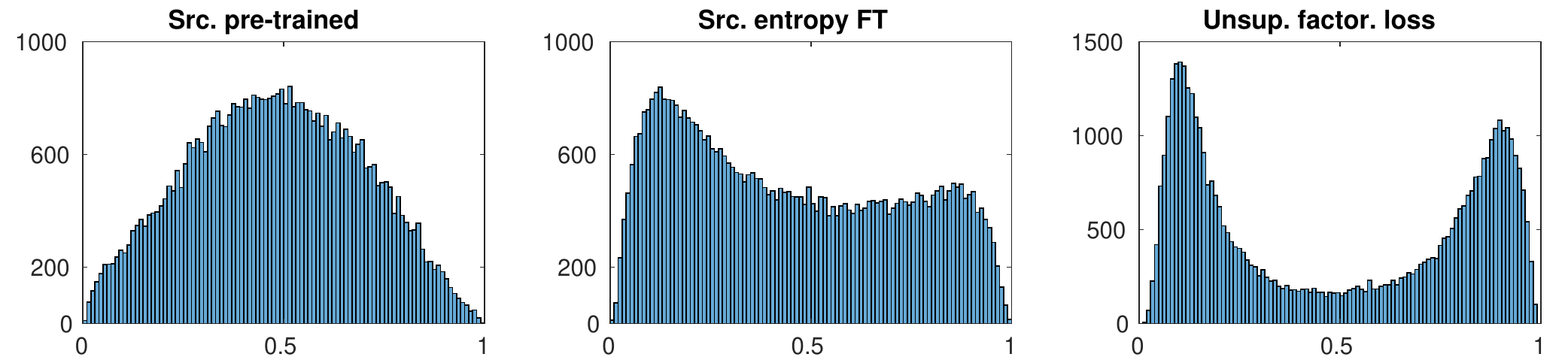}
\caption{CFS activations distribution on target data. Left: Train on source with supervised loss. Middle: Train on source with both supervised and low-entropy CFS losses. Right: CFSM, jointly trained on source and target.}
\label{fig:CFS_distribution}
\end{figure}

\keypoint{Visualisation analysis} To understand the impact of unsupervised factorisation loss, Figure~\ref{fig:CFS_distribution} illustrates the distribution of target CFS activations in the semi-supervised DLSTL setting (SVHN$\to$MNIST). The left plot shows the activations without any such loss, leading to a distribution of moderate predictions peaked around $0.5$. In contrast, the right plot shows the activation distribution on the target dataset of CFSM. We can see that our regulariser has indeed induced the target dataset to represent images with a low--entropy near-binary code. We also compare training a source model by adding low-entropy CFS loss, and then applying it to the target data.
This leads to a low-entropy representation of the source data, but the middle plot shows that when transferred to the target dataset or adaptation the representation becomes high-entropy. That is, joint training with our losses is crucial to drive the adaptation that allows target dataset to be represented with near-binary latent factor codes.

\begin{figure}[t]
\centering    
\includegraphics[width=0.9\columnwidth]{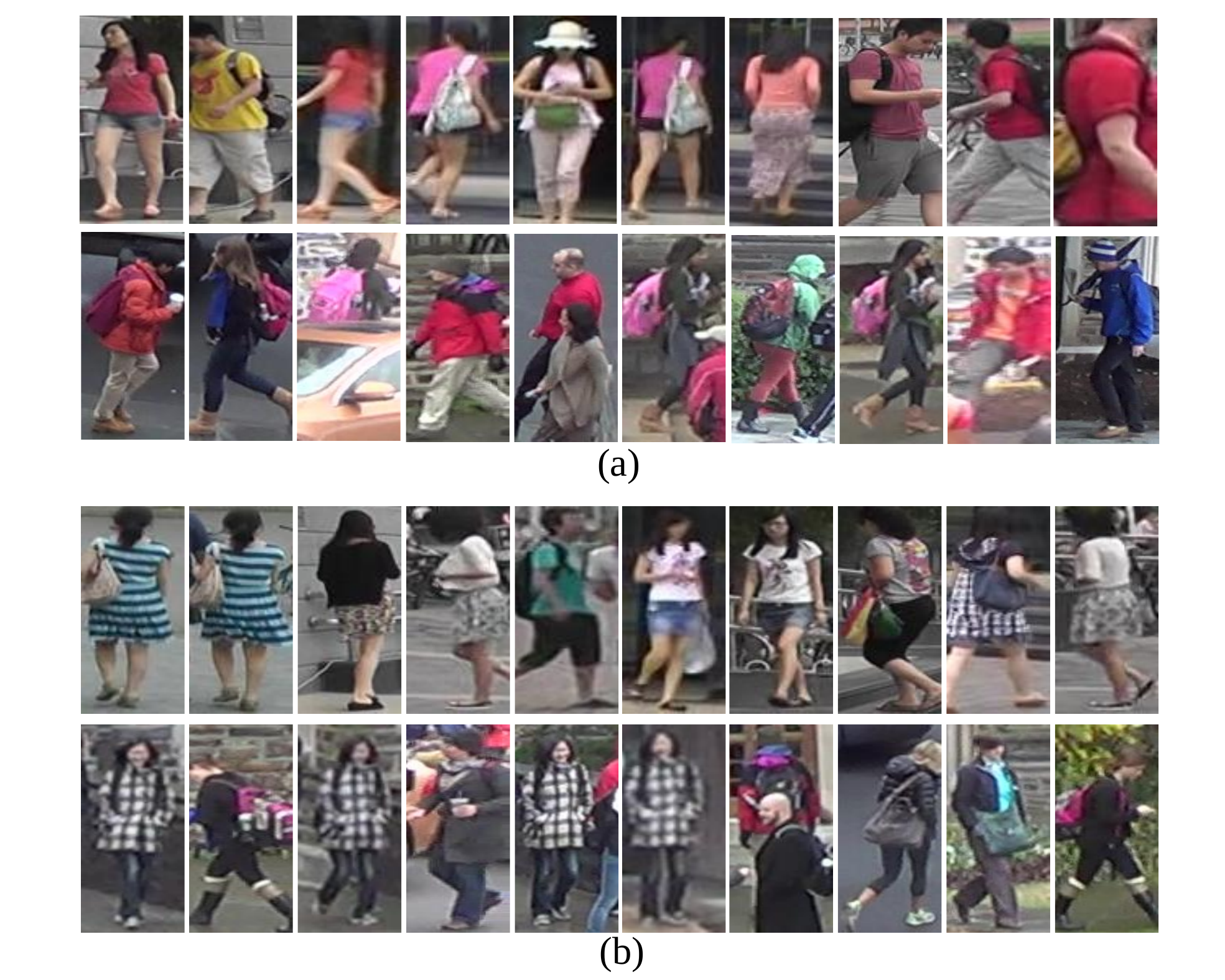}
\caption{Illustration of images selected by two different latent factors: (a) red and (b) female/textured/bag-carrying. In each case the top row is the source (Market) data  and the bottom row is the target (Duke) data. Best viewed in colour.}
\label{fig:CFS_latent_attri}
\end{figure}

\keypoint{Qualitative Analysis} We visualise the discovered latent attributes qualitatively. For each element in $F_C$, we rank images in both source and target domains by their activation. Person images corresponding to the highest ten values of a specific $F_C$ are recorded. Figure~\ref{fig:CFS_latent_attri} shows two example factors with images from the source (first row) and target (second row) dataset. We can see that the first example in Figure~\ref{fig:CFS_latent_attri}(a) is a latent attribute for the colour `red' covering both people's bags and clothes. The second example in Figure~\ref{fig:CFS_latent_attri}(b) is a higher-level latent attribute that is selective for both females, as well as textured clothes and bag-carrying. Importantly, these factors have become selective for the same latent factors across  datasets, although the target dataset has no supervision (i.e., unsupervised DLSTL).


\section{Conclusion}\label{Sec:Con}
We studied a challenging transfer learning setting DLSTL, where the label space between source and target labels are disjoint, and the target dataset has few or no labels. In order to transfer the discriminative cues from the labelled source to the target, we propose a simple yet effective model which uses an unsupervised factorisation loss to discover a common set of discriminative latent factors between source and target datasets. And to improve feature learning for subsequent tasks such as retrieval, a novel graph-based loss is further proposed. Our method is both the first solution to the unsupervised DLSTL, and also uniquely provides a single framework that is effective at both unsupervised and semi-supervised DLSTL as well as the standard UDA. 

\keypoint{Acknowledgements}This work was supported by the EPSRC grant EP/R026173.

{\small
\bibliographystyle{aaai}
\bibliography{egbib}
}

\end{document}